\documentclass[sn-mathphys-num]{sn-jnl}

\usepackage{graphicx}%
\usepackage{multirow}%
\usepackage{amsmath,amssymb,amsfonts}%
\usepackage{amsthm}%
\usepackage{cleveref}
\usepackage{mathrsfs}%
\usepackage[title]{appendix}%
\usepackage{xcolor}%
\usepackage{textcomp}%
\usepackage{manyfoot}%
\usepackage{booktabs}%
\usepackage{algorithm}%
\usepackage{algorithmicx}%
\usepackage{algpseudocode}%
\usepackage{listings}%
\usepackage[italicComments=true, indLines=true, commentColor=gray, spaceRequire=false]{algpseudocodex}
\usepackage{siunitx}
\usepackage{xspace}

\usepackage{svg}
\usepackage{comment}
\usepackage{subcaption}

\usepackage{bbm}
\newcommand{\codofuzz}{\textsf{{CoDoFuzz}}\xspace}

\begin{document}

\title{Robust Black-box Testing of Deep Neural
Networks using Co-Domain Coverage}


\author[1]{\fnm{Aishwarya} \sur{Gupta}}\email{\{aishwaryag, isaha, piyush\}@cse.iitk.ac.in}
\author[1]{\fnm{Indranil} \sur{Saha}}
\author[1]{\fnm{Piyush} \sur{Rai}}

\affil[1]{\orgdiv{Computer Science and Engineering Department}, \orgname{IIT Kanpur}, 
\orgaddress{\country{India}}}

\abstract{Rigorous testing of machine learning models is necessary for trustworthy deployments. We present a novel black-box approach for generating test-suites for robust testing of deep neural networks (DNNs). Most existing methods create test inputs based on maximizing some ``coverage'' criterion/metric such as fraction of neurons activated by the test inputs.  Such approaches, however, can only analyze each neuron's behavior or each layer's output in isolation.  
and are unable to capture their \emph{collective} effect on the DNN's output, resulting in test suites that often do not capture the various failure modes of the DNN adequately. These approaches also require a \emph{white-box} access, i.e., access to the DNN's internals (node activations). We present a novel black-box coverage criterion called \emph{Co-Domain Coverage (CDC)}, which is defined as a function of the model's output and thus takes into account its end-to-end behavior. Subsequently, we develop a new fuzz testing procedure named \codofuzz, which uses CDC to guide the fuzzing process to generate a test suite for a DNN. We extensively compare the test suite generated by \codofuzz with those generated using several state-of-the-art coverage-based fuzz testing methods for the DNNs trained on six publicly available datasets. Experimental results establish the efficiency and efficacy of \codofuzz in generating the largest number of misclassified inputs and the inputs for which the model lacks confidence in its decision. The code is publicly available at \url{https://github.com/aishgupta/Co-domain-coverage.git}.
}


\maketitle

\section{Introduction}

Thorough evaluation of deep neural networks (DNNs) is key for their trustworthy deployment. This requires test suites that potentially cover diverse critical inputs 
and can give us a realistic measure of the DNN's test-set performance to understand and build trust in its functioning, and more importantly to understand what it knows and what it does not. Fuzz testing or fuzzing is a popular approach used in software testing which starts with an initial set of seed test inputs and fuzzes them to generate a richer and more diverse set of test inputs. The fuzzing procedure is typically guided by a \emph{coverage} criterion with the goal to find new inputs that either reveal the software's shortcomings or enhance trust in its capabilities. Such inputs can also help identify the software's weaknesses and improve its performance through further training on these inputs. Motivated by the success of fuzz testing for software systems, recent work has explored this idea for DNN test suite generation as well~\citep{xie2019deephunter, guo2018dlfuzz, kim2019guiding}. The approach is appealing for DNNs because, given the high-dimensional input space, testing the model on every valid input is infeasible.

The outcome of coverage-guided fuzzing depends significantly on the coverage criterion (metric) used during the process. Defining a coverage criterion for DNNs is especially challenging because of the large size of its design space. Most of the several existing coverage criteria~\citep{pei2017deepxplore, ma2018deepgauge} focus on the constituent neurons of the DNN model. The coverage is defined in terms of the activation values of the individual neurons of the DNN and is, consequently, unable to capture the interdependence or correlation among the neurons, and their collective influence on the model's output. Apart from neuron-level coverage criteria, layer-wise coverage criteria~\citep{kim2019guiding, odena2019tensorfuzz, yuan2023revisiting} leverage the output of multiple layers in the DNN model to guide the fuzzing. However, DNNs consist of a large number of layers, 
and applying layer-wise coverage criteria to all the layers can be computationally intensive (in terms of time and memory), constraining such metrics to a subset of layers selected through validation or using some prior knowledge.

An ideal coverage criteria should prioritize test inputs that reveal distinct erroneous behavior of the model and thus its objective to be optimized should be aligned with the end-to-end behavior of the DNN rather than being based on the behavior of its continuent neurons or individual layers. This is especially desirable in cases where we only have access to the DNN as a \emph{blackbox}, i.e., have access to only its predictions for a given input. Motivated by this observation, we propose a novel coverage metric, named \emph{Co-Domain Coverage (CDC)}, which is defined as a function of DNN's output space (a.k.a. its \emph{co-domain}, i.e., a set of all possible outputs that can be predicted by the model). Co-Domain Coverage uses the DNN model as a black box and explores its non-linear, hard-to-interpret mapping of the input space to the output space to generate test-suite inputs. Specifically, Co-Domain Coverage hunts for inputs that can result in diverse outputs; maximizing the number of explored distinct input-output pairs. Moreover, Co-Domain Coverage can guide the fuzzing of the input space, selecting inputs resulting in an output not explored yet. We call this co-domain guided fuzzing as \emph{\codofuzz} which aims at generating a test suite with inputs having varied outputs. 

We evaluate and extensively compare our approach with multiple existing coverage metrics like neuron activation-based coverage namely NC~\citep{tian2018deeptest}, NBC, SNAC, KMNC, TKNP, TKNC~\citep{ma2018deepgauge}, surprise-adequacy guided coverage namely LSC and DSC~\citep{kim2019guiding}, distribution aware coverage NLC~\citep{yuan2023revisiting}, and cluster-based coverage called Tensorfuzz~\citep{odena2019tensorfuzz}. We experiment on six publicly available datasets namely MNIST~\citep{lecun-mnisthandwrittendigit-2010}, FashionMNIST~\citep{xiao2017/online}, CIFAR-10, CIFAR-100~\citep{krizhevsky2009learning}, SVHN~\citep{netzer2011reading}, and Imagenet~\citep{ILSVRC15} and train DNN models of varied sizes and depths according to the complexity of the training dataset. The Co-Domain Coverage guided \emph{\codofuzz} surpasses all the baseline coverage criteria, yielding a test suite with the highest count of erroneous inputs and unearthing the maximum number of distinct errors. It results in a test suite characterized by inputs exhibiting high predictive uncertainty, and diverse class predictions. Moreover, retraining the DNN model on the generated test suite enhances its performance.

\section{Co-Domain Coverage Guided Test Suite Generation}

A DNN represents a function $\mathcal{F}$ mapping high-dimensional inputs $x \in \mathbb{R}^{D}$ to a low-dimensional space $\mathbb{R}^{N}$, where $N$ is the dimensionality of the output. Assuming a classification task, if there are $N$ classes namely $c_1, c_2, \dots, c_N$, and if $c_k$ happens to be the most likely class (i.e., the class with the largest predicted probability $p_{c_k}$) for an input $x$ then the output of the DNN is a tuple $(c_k, p_{c_k})$ denoting the most likely class and its probability. In this case, the co-domain of the DNN is the set of all such possible tuples over all possible inputs.


\begin{figure}[t]
    \centering
    \vspace{-0.4cm}
    \begin{subfigure}{\textwidth}
        \centering
        \includegraphics[width=0.33\linewidth]{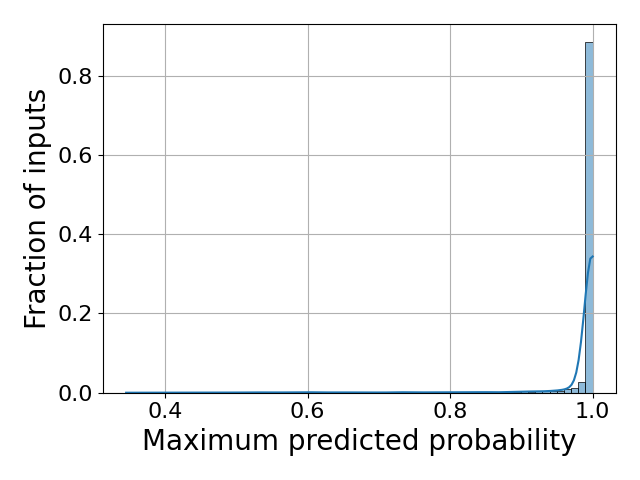}%
        \hfill
        \includegraphics[width=0.33\linewidth]{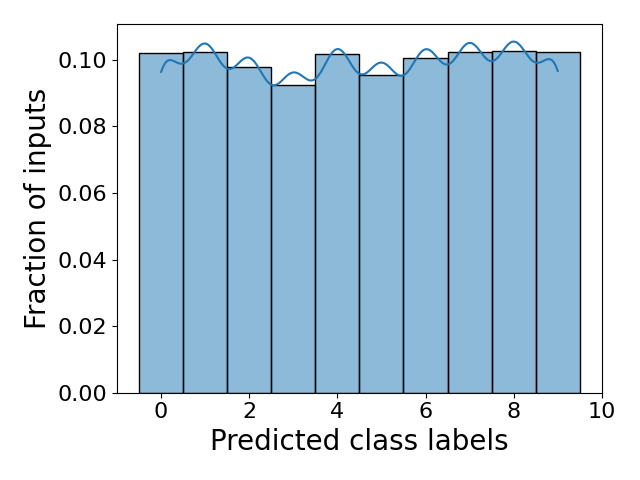}
        \hfill
        \includegraphics[width=0.33\linewidth]{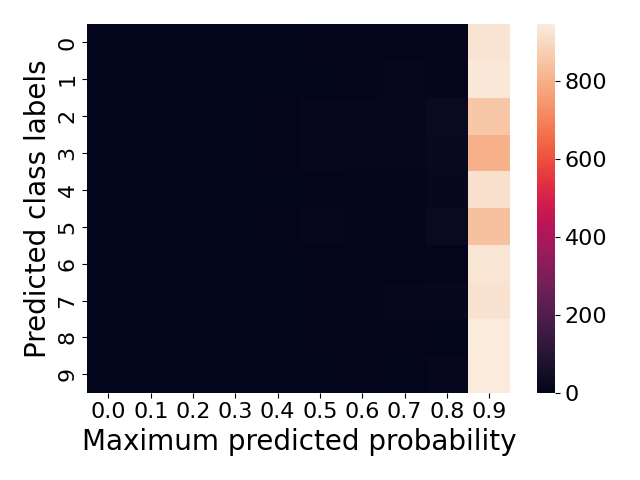}
        \caption{Correctly-classified inputs}
    \end{subfigure}
    \vskip\baselineskip
    \begin{subfigure}{\textwidth}
        \centering
        \includegraphics[width=0.33\linewidth]{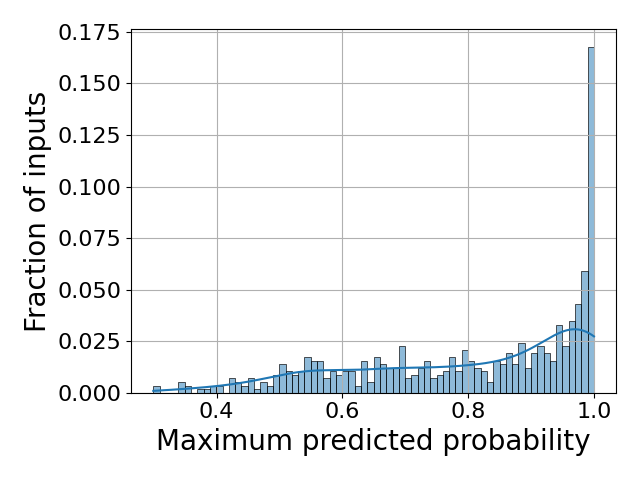}%
        \hfill
        \includegraphics[width=0.33\linewidth]{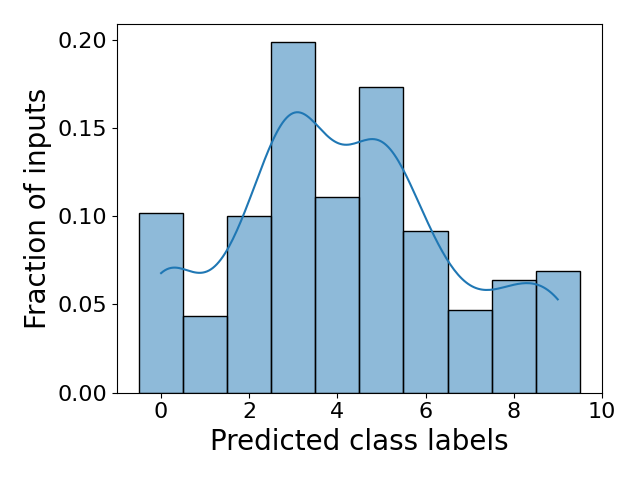}
        \hfill
        \includegraphics[width=0.33\linewidth]{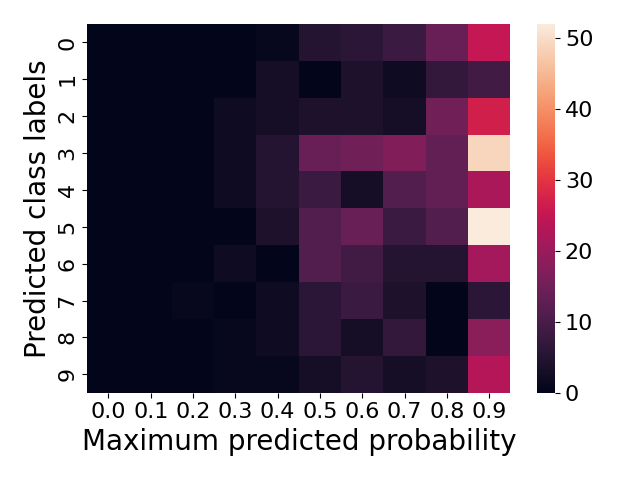}
        \caption{Misclassified inputs.}
    \end{subfigure}
    \caption{\small{Distribution of maximum predicted probability, predicted classes, and discretized co-domain space.}}
    \label{fig:cdc_motivation}
    \vspace{-0.4cm}
\end{figure}

Our method \emph{\codofuzz} is based on analyzing the co-domain, which is easy to interpret and explore and can guide fuzzing to create a test suite of inputs leading to diverse outputs. \emph{\codofuzz}  is based on the observation that the correctly classified and misclassified inputs differ substantially in their output distribution and we will leverage this to design a coverage criterion to identify erroneous inputs, i.e., inputs for which the DNN produces erroneous outputs.
To illustrate this observation, we train a ResNet-18 architecture~\citep{he2016deep} on the CIFAR-10 dataset~\citep{krizhevsky2009learning} achieving an accuracy of $94.22\%$ on its test split, with a reasonable calibration score~\citep{guo2017calibration}. We perform a forward pass on the test split (comprising 10K images not seen during training) of CIFAR-10 and store the output tuple (class label and probability of the predicted class) for all the images in the test split. We then partition the images along with their output tuple into two disjoint sets - one containing all the correctly classified inputs and the other being the set of all the misclassified inputs. 
Next, we visualize the distribution of the probability of the predicted class (i.e., the maximum predicted probability), and the distribution of all the classes in predicted class labels for both the sets of correctly and misclassified inputs separately. The plots are shown in Figure~\ref{fig:cdc_motivation} with the first row comprising plots of correctly classified inputs and the second row corresponding to the misclassified inputs.

The histogram plots show that a large fraction of correctly classified inputs are predicted with high probability (see Figure~\ref{fig:cdc_motivation}~(a)). However, the misclassified inputs exhibit varied probabilities, with the majority having low scores. Furthermore, the distribution of predicted classes is almost uniform for correctly classified inputs (Figure~\ref{fig:cdc_motivation}(b)) but a non-uniform distribution for misclassified inputs i.e., some classes are likely to have larger false positives than others. For instance, misclassified images are more likely predicted to be an instance of class label 3 than that of class label 1. 
Thus, based on these observations, we corroborate that the outputs for misclassified images are widely spread in the co-domain of the network, which needs to be explored exhaustively to find critical inputs.
This motivates the introduction of a novel black-box coverage criterion named \emph{Co-Domain Coverage} leveraging DNN's co-domain to guide the fuzzing of the input space of the DNN model by leveraging its co-domain.

\subsection{Co-Domain Coverage (CDC)}

\label{sec:cdc}

\begin{figure}[t]
\centering
    \includegraphics[scale=0.45]{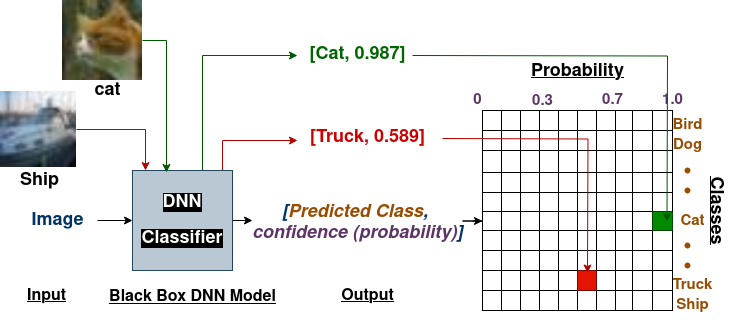}
    \caption{\small{Co-Domain Coverage: Images are given to the black-box DNN model as an input which outputs a tuple of predicted class and its probability. Based on the output tuples, inputs are mapped to a cell in the co-domain of the DNN.}}
    \label{fig:cdc_vis}
    \vspace{-0.4cm}
\end{figure}

Given a set of $K$ inputs $\{x_i\}_{i=1}^K$ and the DNN's predictions in form of their corresponding output tuples $\{(y_i,p_{y_i})\}_{i=1}^K $, we can define their co-domain coverage (CDC) using a two-dimensional matrix (see Fig.~\ref{fig:cdc_vis}) which can be thought of as a clustering of the inputs based on the output tuples. We perform this clustering using a two-level clustering process: first perform a predicted class-based clustering using $y_i$ and then perform class probability-based clustering using $p_{y_i}$. Since $p_{y_i}$ is the probability of the predicted class $y_i$, it always lies in a fixed interval of $[0, 1]$. Therefore, after grouping inputs based on their predicted class, they are further clustered into $M$ clusters based on their maximum predicted probability value. This is done by discretizing the probability range of ($[0, 1]$) into $M$ bins of equal size and mapping the probability of the input to one of these bins. This can be interpreted as binning the inputs into $M$ bins along the real line, with $1/M$ as the length of each bin and $\{i/M\}_{i=0}^{M}$ as the $M+1$ bin boundaries. Furthermore, depending on the size of a bin, we can consider a maximum of $k$ inputs per bin to capture variation among inputs belonging to the same bin (which we refer to as a cluster). Thus, this two-level clustering has $N$ clusters, one per class at the initial level followed by $M$ clusters within each class-based cluster.

The two-level clustering of inputs can be easily implemented by mapping their output tuple to a cell in a 2D matrix of size $(N \times M)$. Each row of this matrix represents a class-based cluster, one row per class. Similarly, the number of columns in every row represents the $M$ value-based clusters for each class. For instance, an input $x_i$ with output tuple $(y_i, p_{y_i})$ will be mapped to the cell $(r, c)$ such that $r=y_i$ is the row index and $c=int(p_{y_i}*M)$ be the column index. Note that the input $x_i$ is mapped to cell $(r, c)$ only if the number of already assigned inputs to this cell is less than the upper bound $k$. The computation of CDC for any input, i.e., mapping it to a discretized co-domain of the DNN is visualized in Figure~\ref{fig:cdc_vis}. For instance, the model outputs the tuple $(truck, 0.689) $ for the image of the class ``ship" (ground truth class). As a result, the image is assigned to the row corresponding to the class ``truck" and is mapped to the bin index $int(0.689*10) = 6$ (with the starting index as 0) in the co-domain matrix. Note that though we select the first $k$ inputs mapped to a cell and ignore the rest, it is also possible to select $k$ most diverse inputs per cell depending on the available compute budget. The pseudocode for computing CDC is summarized in Algorithm~\ref{cc_algo}. 

\begin{algorithm}[t]
    \caption{Co-Domain Coverage}
    \label{cc_algo}
    \begin{algorithmic}
    \Require DNN model $\mathcal{F}$ for an $N$-class classification problem
    \Require Number of bins $M$
    \Require Maximum inputs per cell $k$
    \Require An input $x$
    \Procedure{init-cov-metric}{$N$, $M$}
        \LComment{$Cov$ is a 2D array where rows and columns represent class-based and value-based clusters respectively}
        \For{$r \in \{0,\ldots,N\}$}
            \For{$c \in \{0,\ldots,M\}$}
                \State $Cov(r,c) \gets 0$
            \EndFor
        \EndFor
        \State \Return $Cov$
    \EndProcedure
    
    \Procedure{update-cov}{$Cov, x$}
        \State $y \gets \mathcal{F}(x)$
        \State $prob \gets \mathtt{softmax}(y)$
        \LComment{Indexing is done based on the predicted class and its probability}
        \State $r \gets argmax(prob)$
        \LComment{\Call{bin}{} maps the input to the correct bin}
        \State $c \gets$ \Call{bin}{$max(prob)$}
        \If{$Cov(r,c) < k$} 
            \State $Cov(r,c) \gets Cov(r,c)+1$
            \State \Return True
        \EndIf
    \State \Return False
    \EndProcedure
    \end{algorithmic}
\end{algorithm}

Since the co-domain has been formalized and can be implemented using a 2D matrix, we visualize the co-domain covered by the outputs of correctly classified and misclassified inputs in the last column of Figure~\ref{fig:cdc_motivation} respectively. A clear distinction is observed between the co-domain covered by these two sets of inputs. The correctly classified inputs mostly cover high-probability regions of the co-domain reflecting high-probability predictions. Furthermore, all rows are covered similarly showing a uniform distribution of predicted classes as observed in subplot (a) of Figure~\ref{fig:cdc_vis}. Similarly, in the covered co-domain of misclassified inputs, a large region of co-domain is covered showing predictions with varied probabilities, and different rows are covered unequally reflecting the skewed distribution of predicted classes (subplot (b) of Figure~\ref{fig:cdc_vis}). Thus, all inferences drawn from the covered co-domain are in perfect alignment with the observed behavior; corroborating the proposed design of the co-domain coverage to cover the output space of the DNN model.

Note that though we use fixed $k$ and the same $M$ for all classes, our approach can also work with different $M$ for different classes, to select different numbers of test inputs for different classes. Similarly, we can have different $k$ for different cells; analogous to assigning varied importance to different cells of the matrix. Furthermore, using two hyperparameters $M, k$ instead of one $(M \cdot k, 1)$ adds extra flexibility to regulate the strictness of the Co-Domain Coverage. Notably, when $k>1$, it upper bounds the maximum number of inputs within any bin but the distribution of outputs within a bin is unrestricted. 




\subsection{Computation of the Coverage of Co-Domain of DNN}
The discretized co-domain of the DNN is represented by a 2D matrix $Cov$ of size $N \times M$. Each occupied cell of the matrix implies a covered region in the co-domain. Also, at most $k$ inputs can be mapped to the same cell to increase the local variance of the cell. Thus, the coverage achieved by CDC on a given set of inputs is computed using matrix $Cov$, representing the covered co-domain space by the inputs. We propose two ways to compute the coverage:
\begin{itemize}
    \item \textbf{cdc:} It is defined as the total number of cells occupied by at least one input divided by the total number of cells in the matrix. We can use the indicator function $\mathbbm{1}$ to count the number of occupied cells in $cov$ and compute the coverage as $\mathtt{cdc} = \frac{\sum_{r=0}^{N-1} \sum_{c=0}^{M-1} \mathbbm{1}[Cov(r,c)>0]}{NM}$.
    
    \item \textbf{$k$-cdc:} It is defined as the total number of inputs mapped to the matrix divided by the maximum number of inputs that can be mapped. It can be computed as $\mathtt{k}\text{-}\mathtt{cdc} = \frac{\sum_{r=0}^{N-1} \sum_{c=0}^{M-1} Cov(r,c)}{NMk}$.
\end{itemize}
Note that some of the cells in the 2D matrix cannot be filled. For instance, for N-class classification, the minimum value of the softmax probability of the predicted class will be greater than or equal to $1/N$. Due to this, cells indexed from $0$ to $(\lfloor M/N \rfloor - 1) $ in each row lie in the infeasible region and cannot be occupied. For instance, if $N=10$ and $M=100$, then cells indexed from 0 to 9 (covering probability values $ \in [0, 0.1)$) in each row are infeasible. However, this does not affect the performance of the CDC during fuzzing or computing coverage as either these cells can be explicitly ignored or the indexing can start from $(\lfloor M/N \rfloor) $ onwards. 

\subsection{\codofuzz}
We use our proposed black box coverage criterion called Co-Domain Coverage to guide the fuzzing of the input space and create a test suite by maximizing the coverage of the co-domain of the DNN model under testing. We refer to this black-box fuzzing of DNN as \codofuzz.
We leverage the existing fuzzing framework~\citep{xie2019deephunter} to get \emph{valid, in-distribution} transformed images by applying label-preserving pixel-level transformations (such as color-jitter, contrast enhancement, Gaussian blur) and/or affine transformations (such as crop, scale, flip, and rotation) on the images contained in the seed set (a small set of correctly classified images). The inclusion of the transformed images in the test suite is guided by CDC.
The black box fuzzing of DNN using \codofuzz can be summarized as: 
\begin{itemize}
    \item Select an input from the seed set and apply a label-preserving transformation to get a new modified input.
    \item If the new modified input is valid (in-distribution) 
    and increases the co-domain coverage, then add the modified input to the seed set and the test suite; otherwise, discard it.
    \item Repeat the above two steps until the fixed budget is exhausted.
\end{itemize}
\section{Related Work}
\label{sec:related_work}
Usually, the performance of a DNN is evaluated on a set of inputs that are not seen during training but are randomly drawn from a distribution similar to the training data. Though this is sufficient to quantify the model's performance on clean data, it cannot guarantee a similar performance for inputs that are valid and slightly differ from the training distribution. Therefore, multiple approaches have been proposed to test DNNs, such as coverage-guided testing (\citep{tian2018deeptest, ma2018deepgauge, xie2019deephunter}), concolic testing (~\citep{sun2018concolic, sun2019deepconcolic}), mutation testing (\citep{wei2022mutation, humbatova2021deepcrime, ma2018deepmutation, hu2019deepmutation++}) to evaluate test set quality, differential testing (\citep{pei2017deepxplore}), combinatorial testing (\citep{chen2019variable, ma2019deepct, Gladisch_2020_CVPR_Workshops, chandrasekaran2021combinatorial}), adversarial testing (~\citep{madry2017towards}) and so on. Since our work is about coverage-guided black-box fuzzing, we limit our discussion to existing coverage criteria, fuzzing frameworks, and varied approaches used for black-box testing of DNNs.


\noindent
\textbf{Coverage Criteria.}
Neuron coverage (NC)~\citep{pei2017deepxplore} is the most fundamental neuron activation-based coverage metric, maximizing the number of neurons activated above a specific threshold. It thresholds the continuous-valued neurons' output to a binary value, ignoring the absolute value of their outputs. To address these limitations, multi-granular neuron activation-based coverage criteria~\citep{ma2018deepgauge} has been introduced. Furthermore, the number of neurons to be considered can be reduced by identifying important neurons ~\citep{gerasimou2020importance, xie2022npc} using approaches like layer relevance propagation~\citep{bach2015pixel}. Apart from neurons' activations, coverage criteria have been defined to capture distinct paths~\citep{wang2019deeppath}, sign-value combinations of neuron activations~\citep{sun2019structural}, or 2-way coverage of neurons' triplets~\citep{sekhon2019towards}. 

Moving from neurons to layers, coverage can be defined as identifying distinct activation patterns (Clustering-based Coverage CC~\citep{odena2019tensorfuzz}, maximizing variance of the distribution fitted to the layer's output (NeuraL Coverage (NLC)~\citep{yuan2023revisiting}) or prioritizing relatively more surprising inputs (Surprise Coverage~\citep{kim2019guiding}). However, these methods often face challenges when applied to DNNs with numerous high-dimensional layers, restricting their applicability to a subset of layers (white box) or solely to the output of last layer.

However, unlike above mentioned coverage criteria, CDC leverages the model's output and is independent of the number of neurons, layers, and the latent space dimensionality. It is thus computationally independent of the DNN's size.

\noindent
\textbf{Fuzzing frameworks.}
Multiple testing frameworks exist to generate test inputs. DeepXplore~\citep{pei2017deepxplore} is the first white-box differential testing framework maximizing neuron coverage and is further extended to DLFuzz~\citep{guo2018dlfuzz}. It generates images by optimizing a joint objective of maximizing neuron coverage and minimizing the consensus among DNNs. DeepFault~\citep{eniser2019deepfault} maximizes activations of suspicious neurons to uncover vulnerabilities. DeepTraversal~\citep{yuan2021enhancing} mutates images gradually using low-dimensional data manifolds. Unlike these, DeepTest~\citep{tian2018deeptest} and DeepHunter~\citep{xie2019deephunter} use pixel-level and affine transformations to generate semantically preserved images. 
Since the percentage of generated valid images is maximum ($\approx97-99\%$) for DeepHunter~\citep{xie2019deephunter}, we use its constrained approach to generate \emph{in-distribution} images during fuzzing.

\noindent
\textbf{Black Box Approaches.}
Black box testing approaches do not use activations or any layer's output to generate or evaluate the efficacy of a test suite. Instead, they treat the DNN as a black box and explore the input space, activations, and intermediate layer's output of the surrogate model, or the output of the DNN under test to prioritize inputs, create a test suite, and/or for evaluating and comparing the efficacy of the test suite.
One such approach is to perturb inputs to evaluate the robustness of the model. Wicker et al. use SIFT (Scale Invariant Feature Transform) features to get the saliency distribution of the input pixels to generate the adversarial inputs for testing~\citep{wicker2018feature}. Similarly, BET~\citep{wang2022bet} detects faults in the decision boundaries of the DNN by adding continuous perturbations to inputs. Another black-box approach is to prioritize and select inputs from a large pool of unlabeled inputs. DeepGini~\citep{feng2020deepgini} and DeepGD~\citep{aghababaeyan2023deepgd} are black-box test prioritization approaches that leverage the output of the DNN model to identify inputs that are highly likely to be misclassified by measuring the gini impurity~\citep{quinlan1986induction} and geometric diversity~\citep{aghababaeyan2023black} values respectively. 
Manifold Combination Coverage (MCC)~\citep{byun2021black} is designed to maximize the coverage of the DNN's input space by mapping its high-dimensional input to a low-dimensional manifold using Variational Autoencoders (VAE)~\citep{doersch2016tutorial}. It selects inputs from a master suite comprising of multiple datasets instead of fuzzing the input space like \codofuzz. On the contrary to these approaches, \codofuzz fuzzes the input space by maximizing the co-domain coverage of the DNN model without requiring any additional assistance or auxiliary models. 
\vspace{-1em}
\section{Evaluation}
\label{sec:exp}
We experiment on various datasets using varied neural network architectures and perform a detailed comparison of our method with state-of-the-art coverage-based fuzzing approaches. 
\subsection{Experimental Setup}
\subsubsection{\textbf{Dataset and Neural Network Architectures}}
In our experiments, we use six publicly available datasets:  

\emph{MNIST}~\citep{lecun-mnisthandwrittendigit-2010} It consists of $28 \times 28$ grayscale images of handwritten digits categorized into one of the 10 classes, each class representing a digit between 0-9. It has 60k training and 10k test images. 

\emph{FashionMNIST}~\citep{xiao2017/online} It is a grayscale image classification dataset consisting of images of apparel annotated with one of the ten ground-truth classes. Like MNIST, the image size is $28 \times 28$ and the dataset is divided into train and test splits of size 60k and 10k respectively.

\emph{SVHN}~\citep{netzer2011reading} It consists of $32 \times 32$ RGB images of house number plates labeled with one of the ten digits (0-9). It involves the recognition of digits in natural scene images and has 73257 images for training and 26032 images for testing.

\emph{CIFAR}~\citep{krizhevsky2009learning} It comprises RGB images of size $32\times 32$ for image classification task. We experiment with both CIFAR-10 and CIFAR-100 consisting of images of distinct objects to be categorized into 10 and 100 classes, respectively. Each dataset has 50k training and 10k test images.

\emph{ImageNet}~\citep{ILSVRC15} It is a large-scale RGB image dataset with 1000 categories and an average image size of $469 \times 387$. It has over a million training images and 50k validation images.

Next, to test a DNN model on these datasets, we follow the experimental setup of \citep{pei2017deepxplore, ma2018deepgauge, kim2019guiding},  and train LeNet~\citep{lecun1998gradient} on MNIST and FashionMNIST, VGG-16~\citep{simonyan2014very} on SVHN, ResNet-18, and ResNet-34~\citep{he2016deep} on CIFAR-10 and CIFAR-100, respectively. We use a pre-trained MobileNet-v2~\citep{sandler2018mobilenetv2} available in Torchvision~\citep{torchvision2016} for the ImageNet dataset. These DNN models are of different depths and capacities and thus help evaluate coverage metrics for varied architectures. 
The details of all models such as the training dataset, number of trainable parameters (\#params), number of neurons (\#neurons), number of layers (\#layers), and accuracy on the test split of the dataset, are summarized in Table~\ref{tab:model_accuracy}.
\begin{table}[t]
    \scriptsize
    \setlength{\tabcolsep}{4.2pt}
    \caption{Specifications of DNN models to be tested.}
    \begin{tabular}{l l r r r r r}
        \toprule
        Dataset & Architecture & \#Params & \#Neurons & \#Layers & \multicolumn{2}{c}{Accuracy(\%)} \\
         & & & & & Top-1 & Top-5 \\
        
        \midrule
        MNIST        & LeNet         & 61.7K  & 236   & 5  & 98.57 & - \\
        FashionMNIST & LeNet         & 61.7K  & 236   & 5  & 90.33 & - \\
        SVHN         & VGG-16        & 134.3M & 12426 & 16 & 95.64 & - \\
        CIFAR-10     & ResNet-18     &  11.1M & 4810  & 21 & 94.22 & - \\
        CIFAR-100    & ResNet-34     &  21.3M & 8612  & 37 & 77.56 & 94.22 \\
        ImageNet     & MobileNet-v2  &   3.5M & 18056 & 53 & 71.87 & 90.29 \\
        \bottomrule
    \end{tabular}
    \label{tab:model_accuracy}
    \vspace{-0.4cm}
\end{table}

\smallskip
\noindent 
\textbf{Baselines} We compare Co-Domain Coverage (CDC) with multiple white-box coverage criteria like multi-granular neuron-activations-based coverage criteria~\citep{ma2018deepgauge} like Neuron Coverage (NC), Neuron Boundary Coverage (NBC), Strong Neuron Activation Coverage (SNAC), $K$-Multisection Neuron Coverage (KMNC), Top-$K$ Neuron Coverage (TKNC), and Top-$k$ Neuron Patterns (TKNP), Clustering-based Coverage (CC) ~\citep{odena2019tensorfuzz}, Likelihood-based Surprise Coverage (LSC), Distance-based Surprise Coverage (DSC)~\citep{kim2019guiding}, and recently introduced distribution-based NeuraL Coverage (NLC)~\citep{yuan2023revisiting}. We do not compare with black box approaches~\citep{feng2020deepgini, aghababaeyan2023deepgd, byun2021black} as they have been proposed primarily for test prioritization, and have not yet been adapted for guiding fuzzing. However, baseline approaches like CC, LSC, DSC, and NLC can be adapted for black box settings by only considering the last layer output. In our experiments, we have considered all the layers for these baseline methods, comparing CDC with a more powerful variant than their black box counterparts. Note that CDC substantially differs from the black box variant of these baselines, as it only requires the model's prediction without needing access to the last layer's output.

Most of the baseline coverage criteria have hyperparameters that control the size and quality of the test suite generated by fuzzing the input space but are difficult to choose. So, we follow the experimental setup of Yuan et al.~\citep{yuan2023revisiting} to select the hyperparameter values of all the baseline coverage criteria and list them in Table~\ref{tab:coverage_hp}. 

\smallskip
\noindent
 \textbf{Seed Set and Fuzzing} For each dataset, the seed set consists of a subset of correctly classified images from its test split. For ImageNet, as the ground truth of test data is not public, we use the validation split to create a class-balanced seed set, and the images are resized to $128\times 128$.
 We randomly select a class-balanced subset of 1000 images; 100 images per class for MNIST, FashionMNIST, SVHN, and CIFAR-10 and 10 images per class for CIFAR-100. For ImageNet, a subset of 100 classes is randomly chosen and 10 images are sampled per class to create the initial seed set.
 We adopt DeepHunter~\citep{xie2019deephunter} to perform coverage-guided fuzzing. We use PyTorch~\citep{paszke2019pytorch} for training and testing the DNNs and utilize pixel and affine transformations from torchvision library~\citep{torchvision2016} for mutating images during fuzzing. These transformations include random crop, auto-contrast, jitter, horizontal flip (excluding MNIST and SVHN), rotation, gaussian blur, and random perspective. The constraints imposed by DeepHunter ensure that the transformed images are close to the initial seed image and remain in-distribution for the model.

Since fuzzing is stochastic both in terms of the seeds selected and the order in which transformations are applied, we repeat each experiment 5 times by selecting 5 different seed sets. We use the same seed set and transformations for all coverage criteria across all datasets during fuzzing to ensure a fair comparison. The fuzzing algorithm is terminated if the number of iterations exceeds 10,000 or the running time exceeds 6 hrs. 

\noindent
\textbf{Hyperparameter Selection for CoDoFuzz}
\codofuzz uses Co-Domain Coverage for fuzzing, which has two hyperparameters: the number of value-based clusters per class ($M$) and the maximum number of inputs per cluster ($k$). 
The hyperparameters $M$ and $k$ do not depend on the output of the neurons or intermediate layers of the DNN model and thus are independent of the DNN architecture to be tested. However, they define an upper bound on the size of the test suite selected during fuzzing, i.e., a test suite can contain a maximum of $NMk$ test inputs for a $N$ class classification task. Thus, for fixed $M$ and $k$, the upper bound on the size of the test suite increases with an increase in classes $N$. 
So, to have the same number of maximum possible test inputs across datasets, we set $M=100$ and $k=10$ for MNIST, FashionMNIST, SVHN, and CIFAR-10 datasets, each having 10 ground truth classes, and use $M=10$ and $k=10$ for CIFAR-100 and ImageNet. 
\begin{table}[t]
    \scriptsize
    \centering
    \caption{Hyperparameters selected for all the coverage criteria.}
    \setlength{\tabcolsep}{7.2pt}
    \begin{tabular}{cccccccc}
        \toprule
        NC & KMNC & TKNC & TKNP & LSC & DSC & CC & CDC  \\
        \cmidrule(lr){2-4}\cmidrule(lr){5-6}
        ($th$) & \multicolumn{3}{c}{($k$)} & \multicolumn{2}{c}{($ub/n$)} &  ($d$) & ($M, K$) \\
        \midrule
        0.75 & 100 & 10 & 10 & 0.1 & 0.002 & 10 & 100,10 ($N=10$) \\
         & & & & & & & 10,10 ($N=100$) \\
        \bottomrule
    \end{tabular}
    \label{tab:coverage_hp}
    \vspace{-0.4cm}
\end{table}

\subsection{Research Questions}
We formulate the following research questions and address them empirically to evaluate the test suite obtained by \codofuzz which eventually establishes the efficacy of CDC.  
\begin{itemize}
    \item \textsf{RQ1}: Does \codofuzz succeed in selecting a multitude of inputs on which the DNN model fails? 
    \item \textsf{RQ2}: Does \codofuzz generate a test suite characterized by inputs having high predictive uncertainty?
    \item \textsf{RQ3}: Does \codofuzz result in a test suite with diverse predicted classes?
    \item \textsf{RQ4}: How well does \codofuzz perform in finding errors of distinct types?
    \item \textsf{RQ5}: Does the model's performance improve by retraining on original training set augmented with the test inputs generated by \codofuzz?
\end{itemize}
Furthermore, we raise another question to evaluate Co-Domain Coverage as a coverage criterion.
\begin{itemize}
    \item \textsf{RQ6}: Is there any correlation between the coverage achieved by the CDC on the given set of inputs and the number of erroneous inputs present in it?   
\end{itemize}

We organize our experimental findings in Tables~\ref{tab:misclassified-count}, \ref{tab:entropy-val}, \ref{tab:output-impartiality} and \ref{tab:error-count}, such that each table follows the same structure. Each table represents a subset of evaluation metrics on which test suites created by the guidance of different coverage criteria are evaluated. The columns of the table are arranged into six groups, one per dataset. Each row represents a coverage criteria (mentioned in the first column) used for fuzzing and reports the average performance with the standard deviation of its corresponding test suite.

\subsubsection{\textbf{RQ1: Erroneous inputs in the test suite.}} 
\begin{table}[t]
      \centering
      \scriptsize
       \caption{\small{Number of misclassified test inputs in the test suite.}}
        \begin{tabular}{l r r r r r r }
            \toprule
            Cov & \multicolumn{1}{c}{MNIST} & \multicolumn{1}{c}{Fashion} & \multicolumn{1}{c}{SVHN} & \multicolumn{1}{c}{CIFAR} & \multicolumn{1}{c}{CIFAR} & \multicolumn{1}{c}{IN} \\
            & & \multicolumn{1}{c}{MNIST} & & \multicolumn{1}{c}{10} & \multicolumn{1}{c}{100} &  \\
            \midrule      
            NC &   1$_{\pm00}$ &   2$_{\pm001}$ & 213$_{\pm017}$ &  62$_{\pm010}$ & 146$_{\pm013}$ & 286$_{\pm015}$ \\
            NBC &  12$_{\pm03}$ &  30$_{\pm002}$ & 280$_{\pm018}$ & 313$_{\pm027}$ & 960$_{\pm076}$ & 2002$_{\pm110}$ \\
            SNAC &   5$_{\pm02}$ &  15$_{\pm002}$ & 269$_{\pm013}$ & 186$_{\pm008}$ & 685$_{\pm043}$ & 1524$_{\pm066}$ \\
            KMNC &  50$_{\pm08}$ & 190$_{\pm015}$ & 348$_{\pm072}$ & 886$_{\pm063}$ & 2035$_{\pm114}$ & 3024$_{\pm085}$ \\
            TKNC &   2$_{\pm00}$ &   3$_{\pm001}$ & 191$_{\pm020}$ &  61$_{\pm010}$ & 176$_{\pm006}$ & 488$_{\pm012}$ \\
            TKNP & \textcolor{red}{842$_{\pm44}$} & \textcolor{red}{2227$_{\pm044}$} & 283$_{\pm038}$ & 856$_{\pm061}$ & 1930$_{\pm027}$ & 2947$_{\pm116}$ \\
            LSC & 612$_{\pm75}$ & 1635$_{\pm130}$ & \textcolor{red}{1992$_{\pm106}$} & 1075$_{\pm024}$ & 2598$_{\pm039}$ & 3013$_{\pm90}$ \\
            DSC & 271$_{\pm94}$ & 383$_{\pm045}$ & 1146$_{\pm136}$ & 1455$_{\pm250}$ & 1023$_{\pm081}$ & 537$_{\pm076}$ \\
            CC & 459$_{\pm24}$ & 1494$_{\pm056}$ & 525$_{\pm023}$ & \textcolor{red}{1756$_{\pm039}$} & \textcolor{red}{2762$_{\pm086}$} & \textcolor{red}{3143$_{\pm139}$} \\
            NLC & 477$_{\pm29}$ & 1230$_{\pm125}$ & 322$_{\pm070}$ & 1021$_{\pm071}$ & 2029$_{\pm123}$ & 2743$_{\pm111}$ \\
            CDC & \textcolor{blue}{1487$_{\pm61}$} & \textcolor{blue}{2384$_{\pm082}$} & \textcolor{blue}{2425$_{\pm066}$} & \textcolor{blue}{2496$_{\pm097}$} & \textcolor{blue}{3704$_{\pm098}$} & \textcolor{blue}{5206$_{\pm063}$} \\
            \bottomrule
        \end{tabular}
    \label{tab:misclassified-count}
    \vspace{-0.5cm}
\end{table}
\begin{table}[t]
        \centering
        \scriptsize
         \caption{\small{Average entropy of the test suite to quantify its uncertainty.}}
        \begin{tabular}{l r r r r r r}
            \toprule
            \multicolumn{1}{c}{Cov} & \multicolumn{1}{c}{MNIST} & \multicolumn{1}{c}{Fashion} & \multicolumn{1}{c}{SVHN} & \multicolumn{1}{c}{CIFAR} & \multicolumn{1}{c}{CIFAR} & \multicolumn{1}{c}{IN} \\
            & & \multicolumn{1}{c}{MNIST} & & \multicolumn{1}{c}{10} & \multicolumn{1}{c}{100} & \\
            \midrule      
            NC & 0.25$_{\pm0.04}$ & 0.45$_{\pm0.11}$ & 0.54$_{\pm0.03}$ & 0.30$_{\pm0.03}$ & 0.45$_{\pm0.03}$ & \textcolor{red}{0.99$_{\pm0.05}$} \\
            NBC & 0.11$_{\pm0.01}$ & 0.17$_{\pm0.01}$ & 0.18$_{\pm0.01}$ & 0.13$_{\pm0.01}$ & 0.43$_{\pm0.02}$ & 0.72$_{\pm0.02}$ \\
            SNAC & 0.09$_{\pm0.01}$ & 0.15$_{\pm0.02}$ & 0.23$_{\pm0.01}$ & 0.11$_{\pm0.00}$ & 0.40$_{\pm0.02}$ & 0.73$_{\pm0.03}$ \\
            KMNC & 0.09$_{\pm0.01}$ & 0.18$_{\pm0.01}$ & 0.05$_{\pm0.01}$ & 0.13$_{\pm0.01}$ & 0.47$_{\pm0.02}$ & 0.81$_{\pm0.02}$ \\
            TKNC & 0.19$_{\pm0.05}$ & 0.51$_{\pm0.13}$ & \textcolor{red}{0.61$_{\pm0.02}$} & 0.20$_{\pm0.02}$ & 0.43$_{\pm0.02}$ & 0.81$_{\pm0.04}$ \\
            TKNP & 0.34$_{\pm0.01}$ & \textcolor{red}{0.52$_{\pm0.01}$} & 0.04$_{\pm0.01}$ & 0.12$_{\pm0.00}$ & 0.47$_{\pm0.01}$ & 0.80$_{\pm0.03}$ \\
            LSC & 0.39$_{\pm0.03}$ & 0.50$_{\pm0.01}$ & 0.27$_{\pm0.04}$ & 0.35$_{\pm0.04}$ & 0.94$_{\pm0.02}$ & 0.82$_{\pm0.02}$ \\
            DSC & \textcolor{red}{0.54$_{\pm0.10}$} & 0.45$_{\pm0.03}$ & 0.17$_{\pm0.03}$ & 0.40$_{\pm0.05}$ & \textcolor{red}{0.97$_{\pm0.10}$} & 0.75$_{\pm0.04}$ \\
            CC & 0.25$_{\pm0.01}$ & 0.44$_{\pm0.01}$ & 0.11$_{\pm0.01}$ & \textcolor{red}{0.62$_{\pm0.01}$} & 0.64$_{\pm0.01}$ & 0.83$_{\pm0.04}$ \\
            NLC & 0.25$_{\pm0.02}$ & 0.31$_{\pm0.02}$ & 0.04$_{\pm0.01}$ & 0.13$_{\pm0.00}$ & 0.48$_{\pm0.01}$ & 0.79$_{\pm0.04}$ \\
            CDC & \textcolor{blue}{0.92$_{\pm0.00}$} & \textcolor{blue}{0.87$_{\pm0.01}$} & \textcolor{blue}{0.91$_{\pm0.02}$} & \textcolor{blue}{0.82$_{\pm0.01}$} & \textcolor{blue}{1.55$_{\pm0.01}$} & \textcolor{blue}{1.46$_{\pm0.02}$} \\
            \bottomrule
        \end{tabular}
        \label{tab:entropy-val}
    \caption*{\scriptsize{(\textcolor{blue}{blue} and \textcolor{red}{red} color highlight the best and second-best value in each column respectively).}}
    \vspace{-0.4cm}
\end{table}


The role of a test suite is crucial for fault detection in DNNs, with the effectiveness often tied to the chosen coverage criterion during fuzzing. A test input is deemed \emph{fault-revealing} if the DNN model misclassifies it, identified through a forward pass on the test input. To evaluate coverage efficiency, we quantify fault-revealing test inputs in the test suite and report the average number of misclassified inputs with standard deviation for each coverage criterion in Table~\ref{tab:misclassified-count}. 

As observed from Table~\ref{tab:misclassified-count}, neuron-activation-based coverage performs suboptimally, especially on the MNIST and FashionMNIST datasets, attributed to the small number of total neurons in the LeNet architecture trained on these datasets. Additionally, the majority of LeNet's layers have neurons comparable to TKNC's threshold value of 10, which further explains its poor performance. In general, layer-wise coverage criteria outperform neuron-activation-based approaches, though none of the baselines exhibit consistent performance across various datasets and DNN architectures. In contrast, \codofuzz consistently outperforms other coverage criteria significantly across all DNNs trained on varied datasets except for FashionMNIST where it performs comparably to TKNP. This shows that by maximizing the CDC, \codofuzz can prioritize selection of a high number of fault-revealing inputs in the test suite.

\subsubsection{\textbf{RQ2: Uncertainty Quantification of Generated Test Suite}} 
Similar to a misclassified test input, a correctly classified input but with low confidence can also be a good test case. Such inputs can provide valuable insights into the failure modes and capability of the model. Therefore, a test suite's effectiveness should also be evaluated on the predictive uncertainty of its test inputs. 

For an $N$-class classification task, the output of a model with a $\mathtt{softmax}$ layer on a given input $x$ is a $N$-dimensional probability vector $p = [p_{c_1}, p_{c_2}, \dots, p_{c_N}]$ which sums to 1. We compute the Shannon entropy of this predicted probability distribution $p$ to compute the uncertainty in the model's decision on input $x$. 
Since a test suite comprises a multitude of inputs with varied uncertainty, we compute the entropy of every test input and average it to measure the uncertainty of the test suite. Thus, the entropy of the test suite having $|\mathcal{T}|$ 
inputs can be computed as $\mathcal{U}(\mathcal{T}) = \frac{1}{|\mathcal{T}|}\sum_{j=1}^{|\mathcal{T}|} E(x_j)$ where $E(x_j) = -\sum_{i=1}^N p_{c_i} \log  p_{c_i}$ is the predictive entropy of $x_j$.
The uncertainty scores of all test suites generated using various coverage metrics for all models are computed and presented in Table~\ref{tab:entropy-val}. As evident from the table, the entropy of the test suite generated using \codofuzz consistently surpasses the uncertainty scores of the test suites generated with other criteria by a huge margin. Notably, this trend is more pronounced with the increasing complexity of datasets and the size of DNNs. This can be attributed to the design of the CDC criterion that prioritizes distinct outputs, resulting in a significant proportion of uncertain inputs in the test suite. 
\subsubsection{\textbf{RQ3: Output Impartiality of the test suite}}
\begin{table*}[t]
    \vspace{-0.2cm}
    \centering
    \caption{\small{Output impartiality of the test measured as the number of distinct classes predicted for inputs in the test suite and the normalized Shannon entropy of the output classes.}} 
    \resizebox{.98\textwidth}{!}{
    \begin{tabular}{l r r r r r r r r r r r r}
    \toprule
    Coverage  & \multicolumn{2}{c}{MNIST} & \multicolumn{2}{c}{FashionMNIST} & \multicolumn{2}{c}{SVHN} & \multicolumn{2}{c}{CIFAR-10} & \multicolumn{2}{c}{CIFAR-100}  & \multicolumn{2}{c}{ImageNet} \\
    \cmidrule(lr){2-3}\cmidrule(lr){4-5}\cmidrule(lr){6-7}\cmidrule(lr){8-9}\cmidrule(lr){10-11} \cmidrule(lr){12-13}
    & \multicolumn{1}{c}{$OI$} & \multicolumn{1}{c}{$\#C$} & \multicolumn{1}{c}{$OI$} & \multicolumn{1}{c}{$\#C$} & \multicolumn{1}{c}{$OI$} & \multicolumn{1}{c}{$\#C$} &\multicolumn{1}{c}{$OI$} & \multicolumn{1}{c}{$\#C$} & \multicolumn{1}{c}{$OI$} & \multicolumn{1}{c}{$\#C$} & \multicolumn{1}{c}{$OI$} & \multicolumn{1}{c}{$\#C$} \\
    \midrule\\
    NC & 0.76$_{\pm0.06}$ &   6$_{\pm 0}$ & 0.75$_{\pm0.09}$ &   6$_{\pm 0}$ & 0.97$_{\pm0.01}$ &  10$_{\pm 0}$ & 0.98$_{\pm0.01}$ &  10$_{\pm 0}$ & 0.93$_{\pm0.00}$ &  97$_{\pm 0}$ & 0.71$_{\pm0.01}$ & 201$_{\pm04}$ \\
    NBC & 0.97$_{\pm0.00}$ &  10$_{\pm 0}$ & 0.97$_{\pm0.01}$ &  10$_{\pm 0}$ & 1.00$_{\pm0.00}$ &  10$_{\pm 0}$ & 0.99$_{\pm0.00}$ &  10$_{\pm 0}$ & 0.98$_{\pm0.00}$ & 100$_{\pm 0}$ & 0.74$_{\pm0.00}$ & 429$_{\pm08}$ \\
    SNAC & 0.97$_{\pm0.01}$ &  10$_{\pm 0}$ & 0.96$_{\pm0.01}$ &  10$_{\pm 0}$ & 0.99$_{\pm0.00}$ &  10$_{\pm 0}$ & 0.99$_{\pm0.00}$ &  10$_{\pm 0}$ & 0.98$_{\pm0.00}$ & 100$_{\pm 0}$ & 0.74$_{\pm0.00}$ & 402$_{\pm09}$ \\
    KMNC & 0.97$_{\pm0.01}$ &  10$_{\pm 0}$ & 0.96$_{\pm0.01}$ &  10$_{\pm 0}$ & 1.00$_{\pm0.00}$ &  10$_{\pm 0}$ & 1.00$_{\pm0.00}$ &  10$_{\pm 0}$ & 0.99$_{\pm0.00}$ & 100$_{\pm 0}$ & 0.75$_{\pm0.00}$ & 452$_{\pm07}$ \\
    TKNC & 0.88$_{\pm0.05}$ &   8$_{\pm 0}$ & 0.72$_{\pm0.11}$ &   5$_{\pm 0}$ & 0.98$_{\pm0.01}$ &  10$_{\pm 0}$ & 0.99$_{\pm0.00}$ &  10$_{\pm 0}$ & 0.95$_{\pm0.00}$ &  99$_{\pm 0}$ & 0.74$_{\pm0.00}$ & 281$_{\pm07}$ \\
    TKNP & 0.99$_{\pm0.00}$ &  10$_{\pm 0}$ & 0.99$_{\pm0.00}$ &  10$_{\pm 0}$ & 1.00$_{\pm0.00}$ &  10$_{\pm 0}$ & 1.00$_{\pm0.00}$ &  10$_{\pm 0}$ & 0.99$_{\pm0.00}$ & 100$_{\pm 0}$ & 0.74$_{\pm0.00}$ & 422$_{\pm08}$ \\
    LSC & 0.98$_{\pm0.01}$ &  10$_{\pm 0}$ & 0.99$_{\pm0.00}$ &  10$_{\pm 0}$ & 0.97$_{\pm0.01}$ &  10$_{\pm 0}$ & 0.83$_{\pm0.02}$ &  10$_{\pm 0}$ & 0.95$_{\pm0.00}$ & 100$_{\pm 0}$ & 0.74$_{\pm0.00}$ & 441$_{\pm06}$ \\
    DSC & 0.96$_{\pm0.02}$ &  10$_{\pm 0}$ & 0.98$_{\pm0.01}$ &  10$_{\pm 0}$ & 0.93$_{\pm0.03}$ &  10$_{\pm 0}$ & 0.87$_{\pm0.06}$ &  10$_{\pm 0}$ & 0.90$_{\pm0.02}$ &  99$_{\pm 0}$ & 0.71$_{\pm0.01}$ & 219$_{\pm10}$ \\
    CC & 0.99$_{\pm0.00}$ &  10$_{\pm 0}$ & 1.00$_{\pm0.00}$ &  10$_{\pm 0}$ & 0.99$_{\pm0.00}$ &  10$_{\pm 0}$ & 0.96$_{\pm0.00}$ &  10$_{\pm 0}$ & 0.99$_{\pm0.00}$ & 100$_{\pm 0}$ & 0.75$_{\pm0.00}$ & 446$_{\pm09}$ \\
    NLC & 0.97$_{\pm0.00}$ &  10$_{\pm 0}$ & 0.98$_{\pm0.00}$ &  10$_{\pm 0}$ & 1.00$_{\pm0.00}$ &  10$_{\pm 0}$ & 1.00$_{\pm0.00}$ &  10$_{\pm 0}$ & 0.99$_{\pm0.00}$ & 100$_{\pm 0}$ & 0.74$_{\pm0.00}$ & 399$_{\pm09}$ \\
    CDC & \textcolor{blue}{1.00$_{\pm0.00}$} &  \textcolor{blue}{10$_{\pm 0}$} & \textcolor{blue}{1.00$_{\pm0.00}$} &  \textcolor{blue}{10$_{\pm 0}$} & \textcolor{blue}{1.00$_{\pm0.00}$} &  \textcolor{blue}{10$_{\pm 0}$} & \textcolor{blue}{1.00$_{\pm0.00}$} &  \textcolor{blue}{10$_{\pm 0}$} & \textcolor{blue}{1.00$_{\pm0.00}$} & \textcolor{blue}{100$_{\pm 0}$} & \textcolor{blue}{0.81$_{\pm0.00}$} & \textcolor{blue}{528$_{\pm16}$} \\
     \bottomrule
    \end{tabular}
    }
    \label{tab:output-impartiality}
    \vspace{-0.4cm}
\end{table*}

The coverage-guided fuzzing starts with a class-balanced seed set. In each iteration, it selects one of the seeds, mutates it using the DeepHunter~\citep{xie2019deephunter} algorithm, and adds to the test suite if it increases coverage. The output of the fuzzing is a test suite comprising inputs selected by the chosen guiding coverage criterion and may not be class-balanced, i.e., the distribution of the predicted classes can be skewed toward a subset of the classes. However, in traditional software testing, a test suite should satisfy the output-uniqueness test selection criteria~\citep{harel2020neuron} which states that the test suite should include inputs with diverse behaviors. Thus, to quantify the output impartiality of the test suites $T$, we use existing output-impartiality metric~\citep{harel2020neuron} $OI$ defined as $OI(T) = \frac{\sum_{k=1}^{N} \hat{p}_{c_k} \log\, \hat{p}_{c_k}}{\log\, N}$
where $c_k$ represents the $k^{th}$ class and $\hat{p}_{c_k}$ is the fraction of test inputs predicted to be an instance of class $c_k$ in the test suite. The output impartiality metric is essentially the Shannon entropy of the predicted class distribution of the test suite normalized by $\log N$, where $N$ is the total number of classes. The $OI$ of all the test suites are reported in the column labeled $OI$ in Table~\ref{tab:output-impartiality} for all the datasets. Furthermore, we also report the number of distinct predicted classes for the test suite and show the result in column \#C of Table~\ref{tab:output-impartiality}. 

As evident from Table~\ref{tab:output-impartiality}, all coverage criteria except NC, and TKNC yield a test suite covering all classes.
This is further corroborated by high $OI$ values achieved for most of the test suites which cover all the classes in the output. However, ImageNet is an interesting exception to this trend. Even though the initial ImageNet seed set has only 100 out of 1000 classes, the resultant test suite, generated by all the coverage criteria, encompasses many more predicted classes and has high output impartiality. 
Notably, this is even more pronounced for the test suite obtained with \codofuzz, credited to its emphasis on achieving distinct output tuples (class label and softmax probability value) within the test suite. 

\subsubsection{\textbf{RQ4: Number of Distinct Errors Detected}}

\begin{table}[t]
    \scriptsize
      \centering
       \caption{\small{Number of distinct errors revealed by a test suite.}}
        \begin{tabular}{l r r r r r r }
            \toprule
            \multicolumn{1}{c}{Cov} & \multicolumn{1}{c}{MNIST} & \multicolumn{1}{c}{Fashion} & \multicolumn{1}{c}{SVHN} & \multicolumn{1}{c}{CIFAR} & \multicolumn{1}{c}{CIFAR} & \multicolumn{1}{c}{IN} \\
            & & \multicolumn{1}{c}{MNIST} & & \multicolumn{1}{c}{10} & \multicolumn{1}{c}{100} & \\
            \midrule
            NC &    1$_{\pm 0}$ &    2$_{\pm 1}$ &   52$_{\pm 2}$ &   28$_{\pm 4}$ &   91$_{\pm02}$ &  191$_{\pm12}$ \\
            NBC &   10$_{\pm 2}$ &   20$_{\pm 2}$ &   58$_{\pm 5}$ &   57$_{\pm 4}$ &  480$_{\pm22}$ &  983$_{\pm35}$ \\
            SNAC &    4$_{\pm 1}$ &   11$_{\pm 1}$ &   55$_{\pm 1}$ &   48$_{\pm 2}$ &  373$_{\pm17}$ &  820$_{\pm17}$ \\
            KMNC &   22$_{\pm 4}$ &   38$_{\pm 1}$ &   46$_{\pm 5}$ &   66$_{\pm 2}$ &  643$_{\pm17}$ & \textcolor{red}{1119$_{\pm16}$} \\
            TKNC &    2$_{\pm 0}$ &    3$_{\pm 1}$ &   52$_{\pm 2}$ &   30$_{\pm 3}$ &  127$_{\pm07}$ &  360$_{\pm16}$ \\
            TKNP &   46$_{\pm 0}$ &   50$_{\pm 1}$ &   41$_{\pm 4}$ &   61$_{\pm 3}$ &  588$_{\pm06}$ & 1029$_{\pm29}$ \\
            LSC &   \textcolor{red}{50$_{\pm 1}$} &   \textcolor{red}{51$_{\pm 1}$} &   \textcolor{red}{66$_{\pm 2}$} &   63$_{\pm 2}$ &  758$_{\pm22}$ & 1060$_{\pm23}$ \\
            DSC &   33$_{\pm 2}$ &   37$_{\pm 3}$ &   52$_{\pm 4}$ &   62$_{\pm 5}$ &  282$_{\pm16}$ &  261$_{\pm26}$ \\
            CC &   48$_{\pm 2}$ &   50$_{\pm 1}$ &   61$_{\pm 2}$ &   \textcolor{blue}{84$_{\pm 1}$} &  \textcolor{red}{891$_{\pm29}$} & 1119$_{\pm44}$ \\
            NLC &   28$_{\pm 2}$ &   41$_{\pm 2}$ &   43$_{\pm 3}$ &   64$_{\pm 2}$ &  594$_{\pm18}$ &  905$_{\pm23}$ \\
            CDC & \textcolor{blue}{57$_{\pm 0}$} &   \textcolor{blue}{55$_{\pm 1}$} &   \textcolor{blue}{72$_{\pm 3}$} &   \textcolor{red}{83$_{\pm 2}$} & \textcolor{blue}{1059$_{\pm25}$} & \textcolor{blue}{1660$_{\pm11}$} \\
            \bottomrule
        \end{tabular}
        \label{tab:error-count}
        \vspace{-0.5cm}
\end{table}
\begin{table}
    \scriptsize
    \centering
    \caption{\small{Performance of the DNN model before and after retraining with test suite crafted using coverage-guided fuzzing on the test inputs selected by random fuzzing.}}
    \begin{tabular}{l c c c c}
    \toprule
    Coverage  & \multicolumn{1}{c}{MNIST} & \multicolumn{1}{c}{FashionMNIST} & \multicolumn{1}{c}{CIFAR-10} & \multicolumn{1}{c}{CIFAR-100} \\
    \midrule
    Initial & 92.16$_{\pm0.51}$ & 77.64$_{\pm0.94}$ & 91.71$_{\pm0.65}$ & 81.58$_{\pm0.92}$ \\
    CC & 98.32$_{\pm0.29}$ & 84.50$_{\pm1.04}$ & \textcolor{blue}{93.76$_{\pm0.56}$} & 84.12$_{\pm0.98}$ \\
    NLC & 97.06$_{\pm0.47}$ & 82.03$_{\pm1.74}$ & 93.27$_{\pm0.90}$ & 83.56$_{\pm1.41}$ \\
    CDC & \textcolor{blue}{98.45$_{\pm0.45}$} & \textcolor{blue}{85.14$_{\pm1.25}$} & 93.70$_{\pm0.59}$ & \textcolor{blue}{84.26$_{\pm1.26}$} \\
    \bottomrule
    \end{tabular}
    \label{tab:retraining}
    \vspace{-0.4cm}
\end{table}
In a multi-class classification task, each input is categorized into one of the possible $N$ classes by the model. Out of $N$ possible predictions for input with ground-truth class $i$, any prediction other than $i$ is a misclassification, resulting in $N-1$ types of error and thus a total of 
$N(N-1)$ possible types of errors for $N$ class classification task. 
An informative test suite, thus, should try to detect as many types of errors as possible. The number of types of errors discovered by a test suite can easily be found by doing a single forward pass on its inputs and counting distinct misclassification errors. So we count the number of distinct errors detected by different test suites generated using all baselines and \codofuzz on all the datasets. The number (averaged across five runs in the nearest integer) of distinct errors revealed by test suites is shown in Table~\ref{tab:error-count}. As observed from the table, \codofuzz consistently discovers the maximum number of distinct errors across all the datasets. 

\begin{figure*}[t]
   \begin{tabular}{c c}
    \includegraphics[scale=0.32]{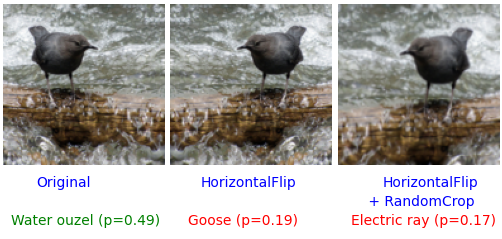}
           &  
    \includegraphics[scale=0.33]{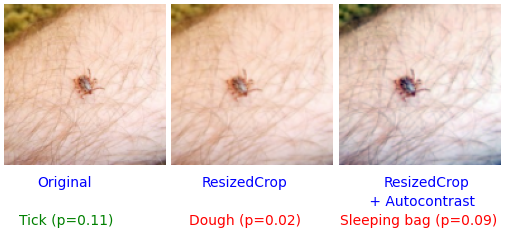}
   \end{tabular}
    \caption{Images from the test suite crafted using \codofuzz. The initial seed images are labeled as original (in blue) with their groundtruth class shown in green color. The model's wrong predictions on the transformed images are displayed in red.}
    \label{fig:in_images}
    \vspace{-0.5cm}
\end{figure*}

We now present some of the images from the test suite crafted using \codofuzz for the ResNet-34 model trained on the ImageNet dataset in Figure~\ref{fig:in_images}. The first image in each set is the original seed image followed by two inputs from the test suite. These test inputs are obtained by applying Torchvision transformations (mentioned in blue color) to the seed image and are misclassified by the model into different classes, illustrating some of the distinct errors unearthed by the test suite. This experiment corroborates the importance of \codofuzz in crafting a test suite enriched with inputs resulting in diverse outputs; unraveling a substantial number of distinct errors possible for the model under testing.
\subsubsection{\textbf{RQ5: Does the model's performance improve by retraining on the original training set augmented with the generated test inputs?}}
 To evaluate the effectiveness of the generated test suite in improving the model's performance, we conduct an experiment by retraining the model on the training dataset augmented with the generated test suite and evaluating its performance. Along with \codofuzz, we consider two strong baselines NLC and CC, and evaluate the efficacy of the generated test suite in enhancing the model's performance. For each coverage metric (namely CC, NLC, and CDC), the model undergoes retraining on a training dataset augmented with the test suite. The experiment is repeated five times, once for each seed set. Subsequently, we create a test suite $T_R$ for these models by randomly fuzzing the input space and limiting the size of test suites to 10K images. We then compare the performance of the DNN model before and after retraining on random fuzz test suite $T_R$. This experiment is repeated and evaluated on four datasets namely MNIST, FashionMNIST, CIFAR-10, and CIFAR-100, and report the averaged results with standard deviation in Table~\ref{tab:retraining}. The first row in Table~\ref{tab:retraining} lists the model's performance on the test suite $T_R$ (created by random fuzzing) before retraining, followed by subsequent rows representing the model's performance after retraining for each coverage criterion. Notably, across three out of four datasets, \codofuzz generated test suite has resulted in maximum improvement in the model's performance on the set $T_R$ after retraining, highlighting its strength and effectiveness.

\subsubsection{\textbf{RQ6: Is Co-Domain Coverage correlated with the number of erroneous inputs}}
\begin{figure}[t]
    \begin{tabular}{c c c c}
         \subfloat[Variation of $\mathtt{Accuracy}$ with the Degree of rotation]
         {\includegraphics[scale=0.22]{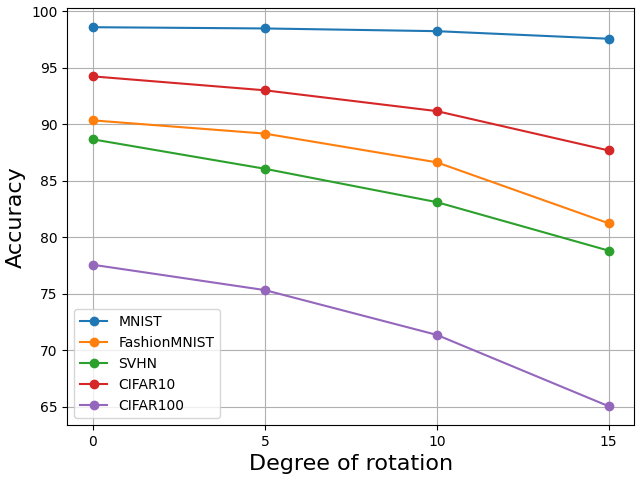}}
         &
         \subfloat[Number of erroneous inputs]{\includegraphics[scale=0.25]{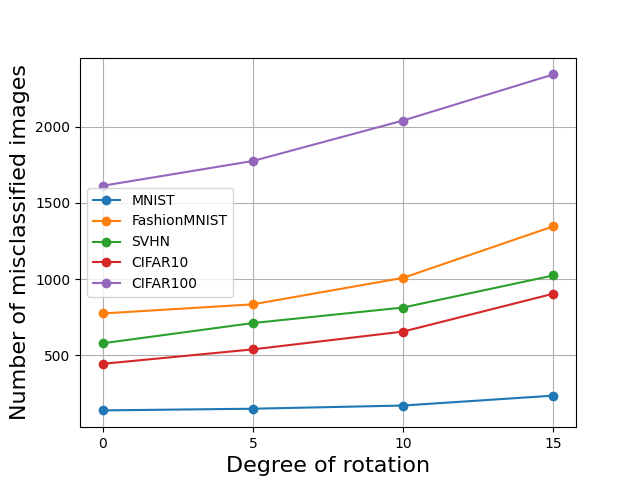}}
         &
         \subfloat[Variation of $\mathtt{CDC}$ with the Degree of rotation]{\includegraphics[scale=0.25]{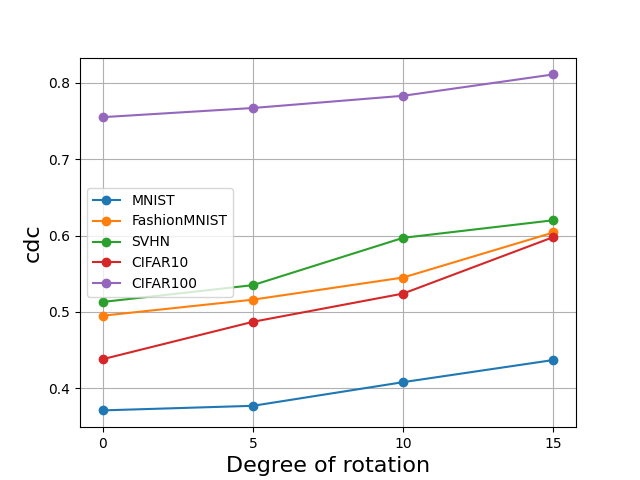}}
    \end{tabular}
    \caption{Correlation between the coverage achieved by CDC and the number of selected erroneous inputs.}
    \label{fig:cdc_correlation}
    \vspace{-0.4cm}
\end{figure}
Given a set of inputs $T$, there should be a correlation between the coverage achieved on $T$ and the number of erroneous inputs present in $T$, i.e., the coverage on $T$ should increase with an increase in number of erroneous inputs in $T$. So we evaluate the correlation between the coverage achieved by the CDC on a given dataset $T$ and the number of erroneous inputs present in $T$. 

We perform this experiment on five datasets and use their original test split $T_0$. The images are rotated in $T_0$ by varying degrees to create datasets with varied accuracy. The images in the original test data $T_0$ are rotated by a maximum of $u^\circ$ for $u^\circ \in \{5^\circ, 10^\circ, 15^\circ\}$ to create a transformed dataset $T_u$. Thus, there are a total of four datasets ($T_0$ and three $T_u$, one per $u$), each with a different number of erroneous inputs. As expected, the accuracy of the DNN model is maximum on $T_0$ (without any rotation) and decreases with the increase in the degree of rotation $u$. The subplot (a) in Figure~\ref{fig:cdc_correlation} illustrates this gradual drop in accuracy with increasing rotation. Notably, the majority of inputs in the rotated datasets are correctly classified, and the number of misclassified inputs increases in the order $T_0 < T_5 < T_{10} < T_{15}$.

Subsequently, we analyze the set of inputs prioritized by the CDC for each $u$. CDC selects a subset of inputs from $T_u$ such that it results in maximum coverage of the co-domain. As $u$ increases, the inputs in $T_u$ are rotated by a larger degree, resulting in more challenging inputs for the model. Thus, a good coverage criterion should be able to select multiple erroneous inputs and the number of selected erroneous inputs should increase with $u$. Thus we count the number of misclassified inputs in the subset of inputs selected by the CDC for all the datasets and $u$ values. We plot the number of erroneous inputs selected by the CDC and achieved coverage for all values of $u$ and on all five datasets in the subplots (b) and (c) of Figure~\ref{fig:cdc_correlation} respectively.
As observed, the number of misclassified inputs selected by the CDC increases with $u$ for all the datasets. Moreover, the coverage achieved by the CDC also increases with $u$ showing a positive correlation between the achieved coverage value and a number of erroneous inputs in the subset of inputs selected by the CDC.
\section{Conclusion}
\label{sec:conclusion}

We have presented a novel coverage criterion for coverage-guided fuzzing of a deep neural network solving a classification problem and evaluated it through a comparison with several state-of-the-art coverage criteria.
The fuzzing method \codofuzz, based on the newly proposed notion of coverage, is highly effective in finding the limitation of DNNs by providing a large number of inputs that a DNN cannot classify correctly or classify with low confidence. Thus, \codofuzz has the potential to be useful in generating effective new training data that will enhance the robustness of a DNN significantly.


\bibliography{references}

\end{document}